\documentclass[runningheads]{llncs}
\usepackage{graphicx}
\usepackage{cite}
\usepackage{amsmath,amssymb,amsfonts}
\usepackage{algorithmic}
\usepackage{graphicx}
\usepackage{textcomp}
\usepackage{array, makecell}
\usepackage{mathrsfs}
\usepackage{amsmath, amssymb,bm}
\usepackage{booktabs}
\usepackage{graphicx,epstopdf,subfigure}
\usepackage{multirow}
\usepackage{tabularx,threeparttable}
\usepackage{array}
\usepackage{xcolor}
\usepackage{comment}
\usepackage{mathtools}
\usepackage{bbding} 
\usepackage{bbm}
\usepackage[ruled]{algorithm2e}

\usepackage{url}  %Required

\usepackage[colorlinks=true,citecolor=blue,urlcolor=blue]{hyperref}

\usepackage{upgreek}

\usepackage{fancyhdr}

\begin{document}
\title{Distribution Aligned Diffusion and Prototype-guided network for Unsupervised Domain Adaptive Segmentation}
\titlerunning{DP-Net}
%\author{Anonymous}
	
	\author{Haipeng Zhou \inst{1}\and
			Lei Zhu\inst{2} \and
			Yuyin Zhou\inst{3}}

	\authorrunning{H.Zhou et al.}
%\authorrunning{Anonymous et al.}
	
	\institute{
	%	School of Computer Science and Technology, 
			Xi’an Jiaotong University, China \\
			\email{lesslie@stu.xjtu.edu.cn}
			\and
%		Robotics and Autonomous Systems Thrust, 
			The Hong Kong University of Science and Technology (Guangzhou), China\and
			University of California, Santa Cruz, USA}
	%
%\institute{Anonymous Organization}
\maketitle              % typeset the header of the contribution

\begin{abstract}
The Diffusion Probabilistic Model (DPM) has emerged as a highly effective generative model in the field of computer vision. Its intermediate latent vectors offer rich semantic information, making it an attractive option for various downstream tasks such as segmentation and detection. In order to explore its potential further, we have taken a step forward and considered a more complex scenario in the medical image domain, specifically, under an unsupervised adaptation condition. To this end, we propose a Diffusion-based and Prototype-guided network (DP-Net) for unsupervised domain adaptive segmentation. Concretely, our DP-Net consists of two stages: 1) Distribution Aligned Diffusion (DADiff), which involves training a domain discriminator to minimize the difference between the intermediate features generated by the DPM, thereby aligning the inter-domain distribution; and 2) Prototype-guided Consistency Learning (PCL), which utilizes feature centroids as prototypes and applies a prototype-guided loss to ensure that the segmentor learns consistent content from both source and target domains. Our approach is evaluated on fundus datasets through a series of experiments, which demonstrate that the performance of the proposed method is reliable and outperforms state-of-the-art methods. Our work presents a promising direction for using DPM in complex medical image scenarios, opening up new possibilities for further research in medical imaging.\footnote[1]{ Code is available at \url{https://github.com/haipengzhou856/DPNet}}
		
\keywords{Unsupervised domain adaptation \and fundus segmentation \and diffusion model}
\end{abstract}
	
\section{Introduction}
% In the glaucoma screening and diagnosis, automated segmentation for the fundus images and privacy protection are the consideration in clinical applications. On the one hand, transplanting a well-trained model in other domains may weaken the performance under the domain-shift problem~\cite{moreno2012unifying}. On the other, the annotations are often inaccessible, especially considering privacy protection in medical image scenarios. 
Automated segmentation of fundus images and privacy protection are important considerations in glaucoma screening and diagnosis. However, 
%in clinical applications, 
there are two significant challenges to address. Firstly, transplanting a well-trained model from one domain to another may lead to a decline in performance due to the domain-shift problem~\cite{moreno2012unifying}.
%as highlighted in~\cite{moreno2012unifying}. 
Secondly, obtaining annotations is often tricky, especially in medical image scenarios where privacy protection is a concern.
% Unsupervised Domain Adaptation (UDA)~\cite{zhang2018task,hoffman2016fcns,javanmardi2018domain,wang2019patch,wang2019boundary,feng2022unsupervised} can solve these two problems very well. The UDA approaches not only align the source and target domain distribution and extract the domain invariants to alleviate the domain gap but save manpower consumption on data labeling.
To address these challenges, unsupervised domain adaptation (UDA)~\cite{zhang2018task,hoffman2016fcns,javanmardi2018domain,wang2019patch,wang2019boundary,feng2022unsupervised} techniques have been developed. UDA approaches aim to align the distribution between the source and target domains, extracting domain-invariant features to alleviate the domain gap. Notably, UDA requires no access to annotations from the target domain, which reduces the manpower consumption needed for data labeling and makes it more practical for medical applications.

%% Generally, there are two mainstreams applied in domain adaptation. One is adversarial learning methods, which usually sets discriminator to leverage the domain distance in input-level~\cite{shaban2019staingan,chen2021source,bateson2020source}, feature-level~\cite{hsu2020every} and output-level~\cite{wang2019patch}. With the discriminator backward the loss of domain discrepancy, the upstream extractor is enforced to learn domain-invariant so that the distribution of different domains is revised and aligned. However, the conventional convolutional neural network (CNN) is relatively weak in feature extraction and distribution alignment. Robust architecture like Transformers~\cite{sun2022safe,yang2023tvt} has indicated that advanced pipelines possess potential in domain generalization and can boost the performance for the downstream work. Recently, Diffusion probabilistic model (DPM) has emerged as a more effective and robust approach to generate images, and it also enlights the downstream works of computer vision like segmentation~\cite{baranchuk2021label}, detection~\cite{chen2022diffusiondet}. However, the research on domain adaptation of DPM is still blank. To the best of our knowledge, we are the first to explore the domain generalization capability of DPM, where we design a domain discriminator to align the inter-domain distribution and obtain the rich semantic representation.

There are two main UDA approaches, namely adversarial learning and intra-domain category rectification. Adversarial learning methods involve training a discriminator to minimize the domain discrepancy between different domains, typically at the input-~\cite{shaban2019staingan,chen2021source,bateson2020source}, feature-~\cite{hsu2020every}, or output-level~\cite{wang2019patch}, enforcing the upstream extractor to learn domain-invariant features. While this approach has shown promise, conventional CNNs are limited in their ability to extract and align features across domains. Advanced architectures such as Transformers~\cite{sun2022safe,yang2023tvt} have shown potential for improving domain generalization. Additionally, Diffusion probabilistic model (DPM) has emerged as a more effective and robust approach for generating images and downstream computer vision tasks such as segmentation~\cite{baranchuk2021label} and detection~\cite{chen2022diffusiondet}. However, their usage in domain adaptation remains unexplored. Another mainstream approach is intra-domain category rectification, which aims to rectify discrepancies.
%within each domain by producing reliable pseudo-labels for the target domain. 
Pseudo-labeling methods~\cite{zheng2021rectifying,yan2022unsupervised,chen2021source} are commonly used for this purpose, but these methods heavily rely on the quality of generated pseudo-labels, which can lead to performance degradation when given uncertain labels. 
% firstly became popular which aims to produce reliable pseudo labels for the target domain doing further optimization, while such kind of approach highly relies on the quality of generated pseudo labels resulting in performance degradation when given uncertain labels. 
To address this issue, uncertainty estimation techniques are widely used to denoise unreliable masks. For instance, DPL~\cite{chen2021source} estimates uncertainty via Monte Carlo Dropout for Bayesian approximation and filters out the noise in the pseudo-label.
% Uncertainty estimation is widely used to denoise the unreliable mask, DPL~\cite{chen2021source} estimates the uncertainty via Monte Carlo Dropout~\cite{gal2016dropout} for Bayesian approximation and filters out the noises in the pseudo label. 
% Though distribution discrepancy is situated in different domains, using domain invariants to make regularization is feasible. ProDA~\cite{zhang2021prototypical} suggests that the prototypes can provide anchors for intra-domain that help extract consistent content. 
Domain invariants, such as prototypes, can also be used to regularize the training process to rectify the distribution discrepancy between different domains. 
% Prototypes are reference representations of features from training samples and are trained to be domain-invariant for consistency across domains. 
ProDA~\cite{zhang2021prototypical} suggests that prototypes can act as anchors for intra-domain category rectification, aiding in extracting consistent content from source and target domains. 
% Aligning prototypes from various domains can improve the model's generalization performance on unseen domains by capturing the data's common structure.
%CLR~\cite{feng2022unsupervised} introduces intra-domain and inter-domain regularization encouraging semantically similar regions to have the same labels. 
% Hence, aligning the prototypes from different domain makes benefits in achieving intra-domain category rectification. 

 In this paper, we investigate the domain generalization capability of DPM and propose a novel Diffusion-based and Prototype-guided network, called DP-Net, for unsupervised domain adaptive segmentation. 
 To the best of our knowledge, this is the first attempt to explore DPM for domain adaptation purposes.
 Our proposed DP-Net comprises two stages: Distribution Aligned Diffusion (DADiff) and Prototype-guided Consistency Learning (PCL) module. To align the inter-domain distribution, the DADiff aims to train a domain discriminator to minimize the difference between the intermediate features of the source and target images generated by the DPM. The PCL stage, on the other hand, promotes consistent learning across different domains by utilizing class-wise feature centroids as domain-invariant prototypes, encouraging the segmentor to learn consistent content. By leveraging the latest diffusion models, DP-Net can extract domain-generalized features more effectively, resulting in superior performance for unsupervised domain adaptive segmentation on fundus image data.
 
% Second, it promotes consistent learning across different domains with the PCL module, where class-wise feature centroids encourage the segmentor to learn consistent content. 
% Lastly, DP-Net utilizes the latest paradigm of diffusion models to achieve superior results.
 
% In this paper, To the best of our knowledge, we are the first to explore the domain generalization capability of DPM, where we design a domain discriminator to align the inter-domain distribution and obtain the rich semantic representation. we propose a Diffusion-based and Prototype-guided network DP-Net for unsupervised domain adaptative segmentation. Our method enjoys the following benefits: 1) extracting the domain generalized features via Distribution Aligned Diffusion (DADiff) resulting in better domain adaptive segmentation; 2) emphasizing consistent learning across different domains with Prototype-guided Consistency Learning (PCL) module such that the class-wise feature centroids courage the segmentor to learn the consistent content; 3)  exploring the up-to-date paradigm of diffusion model which can obtain preferable results.
%4) experiments and ablation analysis suggests that our framework is effective and outperforms other state-of-the-art methods on the public fundus image dataset.

\section{Method}
Our proposed DP-Net comprises two stages. In stage 1, we introduce the Distribution Aligned Diffusion (DADiff) model that combines a diffusion model with a domain discriminator to extract generalized features from both source and target images. DADiff mitigates the domain gap and yields latent representations, denoted as $F^S$ and $F^T$ (Fig.~\ref{stage1}). In stage 2, we design a Prototype-guided Consistency Learning (PCL) module, where $F^S$ is used to train $G$ to predict the source domain's segmentation map. We generate a pseudo-label with $F^T$ and compute a prototype-based consistency loss, incorporating both domain segmentation outcomes to guide $G$ to prioritize consistent objectives (Fig.~\ref{stage2}).
% . This consistency loss guides $G$ to prioritize consistent objectives, as illustrated in Fig.~\ref{stage2}.

% Our Diffusion-based and Prototype-guided network (DP-Net) has two stages. At stage 1, 
% %we devise a Distribution Aligned Diffusion (DADiff) by combining the diffusion model and a domain discriminator to extract generalized features, as shown in  Fig.~\ref{stage1}. 
% by passing the source and target images into our DADiff, we can alleviate the domain gap via a domain discriminator and obtain the latent representations (see $F^S$ and $F^T$) from the diffusion network.
% %
% For stage 2, we input the features $F^S$ from the source domain to a segmentor $G$ to predict the segmentation map and compute a supervised loss. Then, we send the features $F^T$ to the segmentor to obtain the pseudo label. 
% %
% After that, we use the segmentation results of the source domain and the target domain to compute a prototype-based consistency loss to further instruct $G$ to focus on the consistent objectives.
%feature pass the extracted features from stage 1 to the Prototype-guided Consistency Learning (PCL) Module (PCLM). By computing the source and target domain prototypes, the prototype-guided consistency learning loss instructs the segmentation module to focus on the consistent objectives.
	
\subsection{Distribution Aligned Diffusion (DADiff) model}
\begin{figure*}[!t]
		\centering
		\includegraphics[width=0.95\textwidth]{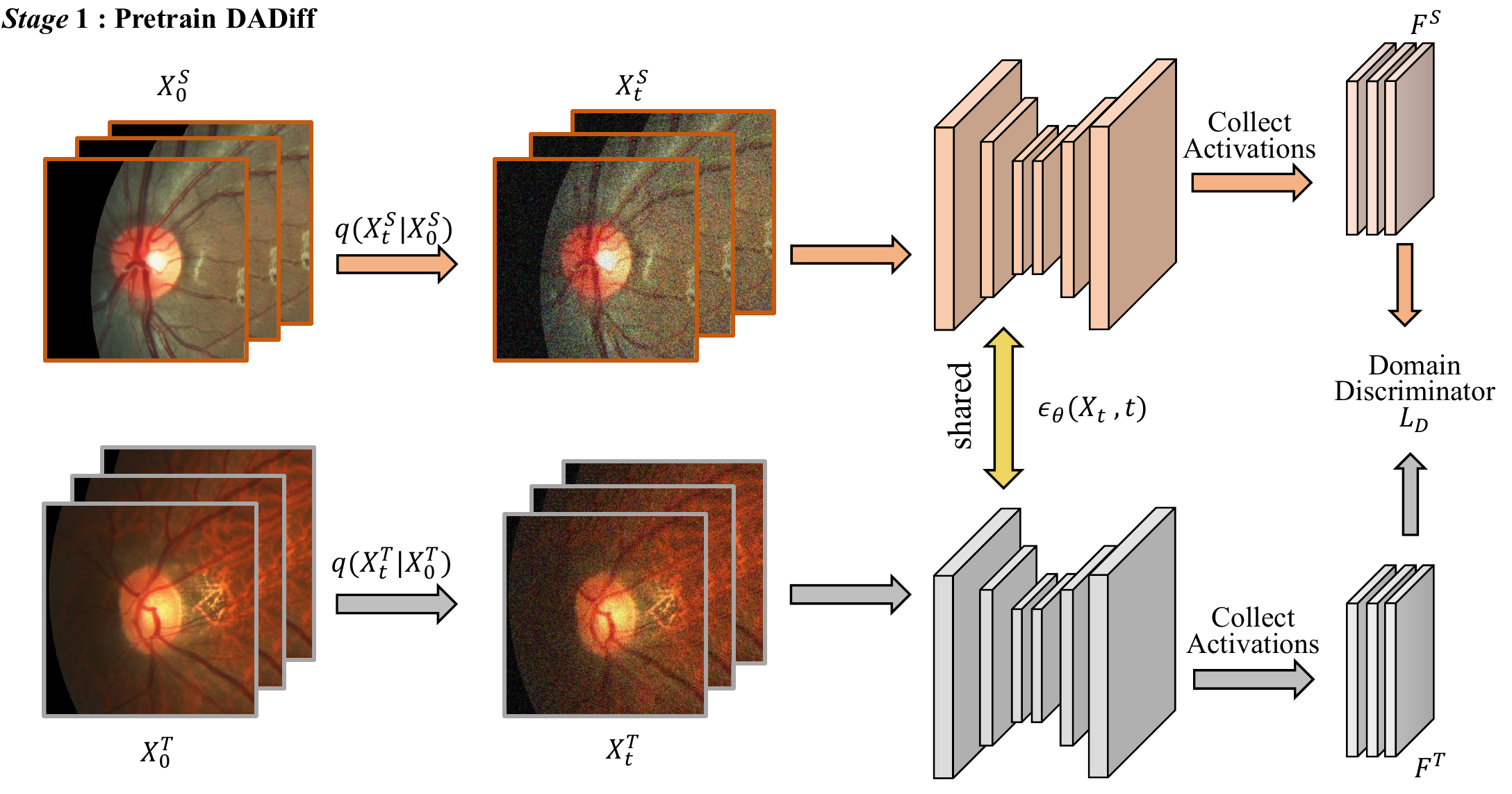}
        \vspace{-3mm}
		\caption{Illustration of our DADiff at stage 1. We deploy the diffusion model and utilize the latent activations to extract diffusion features from different domains. And then we devise a domain discriminator on the diffusion features to reduce their domain gaps.}
		\label{stage1}
\end{figure*}

%\begin{equation} \label{eq2}
%q(x_{t}|x_{0})=\mathcal{N}(x_{t};\sqrt{\bar{\alpha_{t}}}x_{0}, (1-\bar{\alpha_{t}})I),
%\end{equation}
The exploration of latent feature representations from generative models has been widely investigated in many dense prediction tasks~\cite{tritrong2021repurposing,xu2021generative}. However, GANs are prone to collapse and out-of-distribution issues. On the other hand, the diffusion model has attracted much research attention and has demonstrated superior performance in many vision tasks due to its noise robustness.

% Exploring the latent feature representation from generative models has been widely investigated in
% many dense prediction tasks~\cite{tritrong2021repurposing,xu2021generative}. 
% Note that GAN may receive collapse and our-of-distribution.
% Diffusion model has attracted much research attention and has achieved superior performance in many vision tasks due to its noise robustness.
%, recently DPM with robustness is better to perform this task. 
% Instead of taking the labels as input in the original DPM~\cite{amit2021segdiff}, we adapt the vanilla Diffusion~\cite{dhariwal2021diffusion} originally used to generate an image. 
% Considering the intermediate latent features contain rich semantic information~\cite{baranchuk2021label}, we discard the produced images and only wrap up the activations of DPM to obtain latent feature representations.
% %%
% Hence, we devise a distribution Aligned Diffusion model (DADiff), see Fig. \ref{stage1} for the schematic illustration of our DADiff model.

%%
Instead of taking labels as input in the original DPM~\cite{amit2021segdiff}, we adapt the vanilla Diffusion~\cite{dhariwal2021diffusion}, which is originally used to generate an image. This allows us to obtain latent feature representations from the intermediate activations of DPM, which contain rich semantic information~\cite{baranchuk2021label}. In order to align the feature distributions between the source and target domains, we propose Distribution Aligned Diffusion (DADiff), as shown in Fig. \ref{stage1}. The generated images are discarded, and only the activations features of DPM are utilized. %to obtain the latent feature representations.
%%	
% The first step of our DADiff is to follow the design of the classical diffusion model to train a U-Net for predicting noise.
% Like the classical diffusion model, our DADiff adds a series of Gaussian noises to the sample in the forward process and restores it to predict the noise in the diffusion reverse process. 
%We follow the classical diffusion model architecture and train a U-Net to predict noise. 
DADiff introduces a sequence of Gaussian noises to the input image in the forward pass and reconstructs the image in the reverse pass to predict the noise.
Specifically, given a sample $x_{0}$, the forward process $q$ is:
\begin{equation}\label{eq1}
	q(x_{t}|x_{t-1})=\mathcal{N} (x_{t}; \sqrt{1-\beta_{t}}x_{t-1},\beta_{t}I),
\end{equation}
where $\beta_{t}$ denotes the variance condition at the step $t$. $I$ is the identity matrix. $\mathcal{N}$ is a distribution. 
The noisy sample $x_{t}$ at the time step $t$ can be obtained as:
	\begin{equation} \label{eq3}
		x_{t} = \sqrt{\bar{\alpha_{t}}}x_{0}+\sqrt{(1-\bar{\alpha_{t}})}\epsilon, \quad \epsilon \sim\mathcal{N}(0,1).
	\end{equation}
where $\alpha_{t}=1-\beta_{t}$. $\bar{\alpha_{t}}=\prod_{n=1}^{t}\alpha_{n}$. 
Meanwhile, the reverse diffusion process $p_{\theta}$ is parametered by $\theta$ and can be described as:
	\begin{equation}\label{eq4}
		p_{\theta}(x_{x-1}|x_{t})=\mathcal{N}(x_{x-1};\mu_{\theta}(x_t,t),\Sigma_{\theta}(x_t,t)).
	\end{equation}
Instead of directly obtaining the distribution $\mathcal{N}$ of Eq.~\ref{eq4}, our DADiff follows~\cite{ho2020denoising} to predict the noise via a UNet $\epsilon_{\theta}$:
	\begin{equation}\label{eq5}
		x_{t-1} = \frac{1}{\sqrt{\alpha_{t}}} (x_{t}-\frac{1-\alpha_{t}}{\sqrt{1-\bar{\alpha_{t}}}}\epsilon_{\theta}(x_{t},t))+\sigma_{t}z, \quad z\sim\mathcal{N}(0,I) ,
	\end{equation}
where $\sigma_{t}$ denotes the variance scheme learnt by $\epsilon_{\theta}$. $\epsilon_{\theta}$ is often assign with a UNet. We follows~\cite{dhariwal2021diffusion} to compute the noise estimation loss to train the UNet $\epsilon_{\theta}$.
%Typically the model adopts U-Net structure, and we set the one proposed in~\cite{dhariwal2021diffusion} following its loss to train the U-Net. 

% We feed the source image $x_{0}^{S}$ into the trained U-Net to extract the corresponding latent diffusion features $F^S$. Similarly, the target image $x_{0}^{T}$ is also processed by the trained U-Net to generate the diffusion features $F^T$
%obtain the and compute a domain discriminator on these features of the image
%%
%Taking the source image $x_{0}^{S}$ and target image $x_{0}^{T}$ as input, our DADiff follow Fig. \ref{stage1} to execute the Markov step and collect the latent activation features $F^{S}$ and $F^{T}$ respectively in the reverse diffusion process. 
% To reduce the domain gap between the source domain and the target domain, we devise a domain discriminator $D$ embedded a gradient reversal layer (GRL)~\cite{ganin2015unsupervised} to make inter-domain indistinguishable. 
We utilize the trained UNet to extract latent diffusion features $F^S$ from the source image $x_{0}^{S}$ and $F^T$ from the target image $x_{0}^{T}$. Specifically, in the reverse diffusion process, our DADiff follows the Markov step to collect the latent activations $F^{S}$ and $F^{T}$, respectively.
To reduce the domain discrepancy between these two domains, we implement a domain discriminator $D$ with a gradient reversal layer (GRL)~\cite{ganin2015unsupervised} to ensure inter-domain indistinguishability, as shown in Fig. \ref{stage1}.
Specifically, we devise the following loss $\mathcal{L}_{D}$ to learn the domain discrimintor:
	\begin{equation}\label{eq6}
		\begin{split}
			\mathcal{L}_{D}=-\sum_{x,y}\left[l^S \log \left(D  \left(F^S \right)^{ \left(x,y\right)}\right)+  \left(1-l^T\right) \log \left(1-D\left(F^T\right)^{\left(x,y\right)}\right)\right] .
		\end{split}
	\end{equation}
where ${l}^S=1$ and ${l}^T=0$ denote the source and target domain labels, respectively. The coordinate in the diffusion feature map is denoted as $(x, y)$.
%With the constraint of the domain discriminator, the latent layers of the U-Net keep restoring the image but under a domain-generalized condition. Such that, we can revise the distribution implicitly and obtain the wrap-up intermediate generalized features.
 
\subsection{Prototype-guided Consistency Learning (PCL) Module}
	\begin{figure*}[!t]
		\centering
		\includegraphics[width=0.95\textwidth]{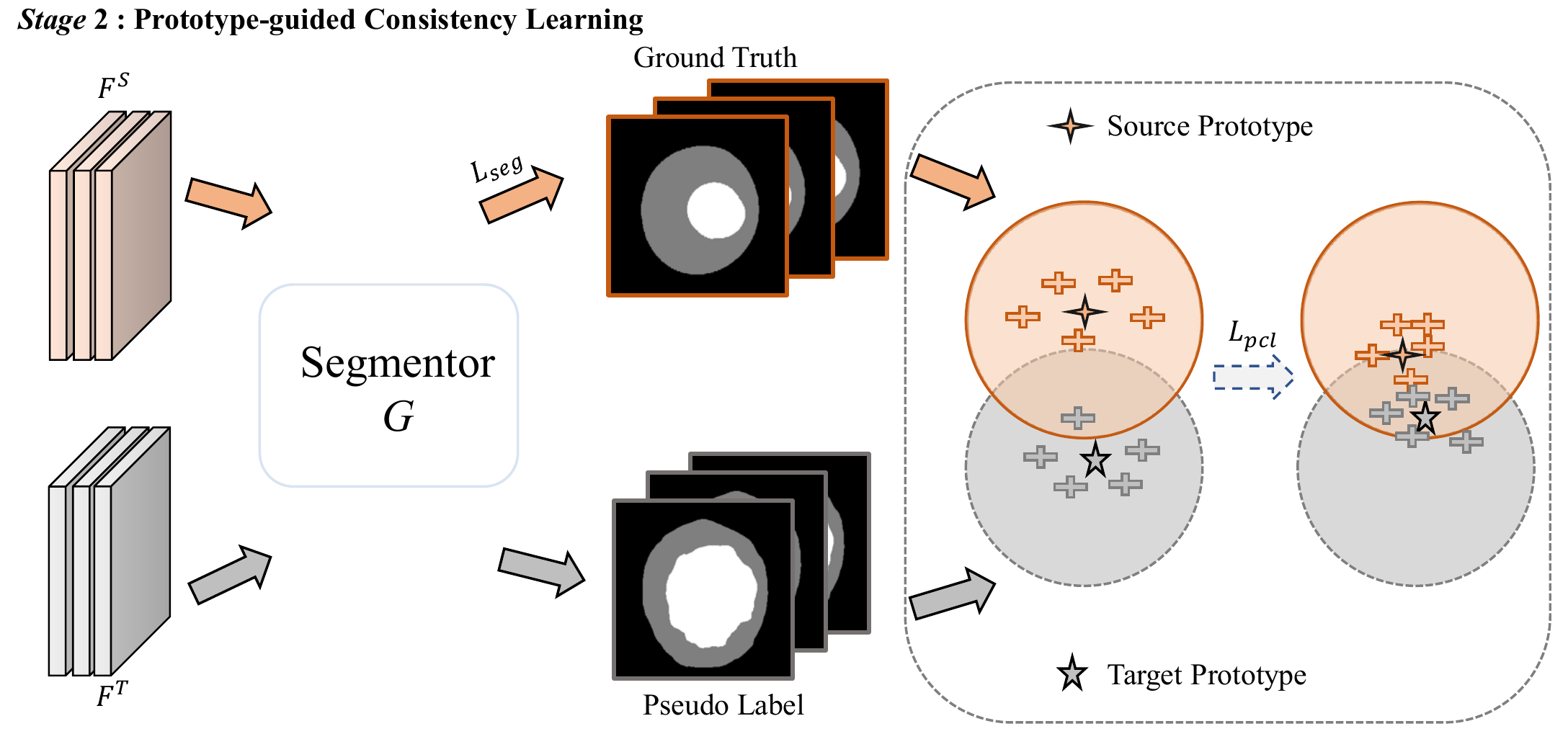}
        \vspace{-5mm}
		\caption{Overview of our stage 2 with a PCL module. Inputting the features from stage 1, we deploy a shared segmentor G to predict segmentation at the source and target domains.
        Then, we compute prototypes of both domains and devise a prototype-guided consistent loss to reduce the prototype distance for domain adaptation.}
		\label{stage2}
	\end{figure*}

Given the latent diffusion features generated by DADiff in the first stage, we introduce a Prototype-guided Consistency Learning (PCL) module in the second stage to further narrow the gap between the source and target domains, where the detail is shown in Fig.~\ref{stage2}.
%Fig.~\ref{stage2} provides a schematic illustration of our PCL module. 
% we devise a Prototype-guided Consistency Learning (PCL) module in stage 2 to further reduce the gap between the source domain and the target domain. 
%further improve the domain adaptation performance.
% Fig.~\ref{stage2} shows the schematic illustration of our PCL module, which takes the latent diffusion features at stage 1 as input. 
% Specifically, as an UDA setting, we have the source domain features $\{F_{i}^{S}\}^{N_S}_{i=1}$ and their segmentation annotations $\{Y_{i}^{S}\}^{N_S}_{i=1}$ and the target domain features $\{F_{i}^{T}\}^{N_T}_{i=1}$ without any label. 
% Then, we pass the source domain features to a shared segmentation decoder $G$ to obtain the segmentation results and then perform supervised learning on the segmentation results at the source domain.
% Similarily, we obtain the pseudo label $\{\hat{Y}_{i}^{T}\}^{N_T}_{i=1}$ for the target domain under a probability threshold selection $\mathbbm{1}[p^{T}\ge\gamma]$. 
In the UDA setting, we have access to the source domain features $\{F_{i}^{S}\}^{N_S}_{i=1}$ and their corresponding segmentation annotations $\{Y_{i}^{S}\}^{N_S}_{i=1}$, as well as the target domain features $\{F_{i}^{T}\}^{N_T}_{i=1}$  without any label. To perform domain adaptation, we first pass the source domain features through a shared segmentation decoder $G$ to obtain segmentation results, on which we perform supervised learning to improve the segmentation performance on the source domain.
Similarly, we generate pseudo labels $\{\hat{Y}_{i}^{T}\}^{N_T}_{i=1}$ for the target domain based on a probability threshold selection $\mathbbm{1}[p^{T}\ge\gamma]$.
%With the last convolution of the decoder, one can obtain the output feature $f^{S}$ and $f^{T}$,
%Then, based on the diffusion feature $f^{S}$ at the source domain and $f^{T}$ at the target domain, we compute the object prototype of the source and target domain as following:
With inputting $F^S$ and $F^T$, one can obtain the output feature $f^{S}$ and $f^{T}$ from the last convolution layer of the segmentation decoder. And we compute the object prototype $z_{obj}^{S}$ at the source domain and the object prototype $z_{obj}^{T}$ at and target domain as:
	\begin{equation}\label{eq7}
		z_{obj}^{S}=\frac{1}{N_{obj}^{S}}\sum_{i}\mathbbm{1}(Y_i^{S}=1)f^S, \quad 	 z_{obj}^{T}=\frac{1}{N_{obj}^{T}}\sum_{i}\mathbbm{1}(\hat{Y}_{i}^{T}=1)f^T,
	\end{equation}
where $i$ denotes the index of pixel and $N_{obj}$ is the number of pixels of object classes. However, the prototype of the target domain may be unreliable due to false predictions produced by $G$. To address it, we follow~\cite{chen2021source} and estimate the uncertainty to denoise the coarse pseudo labels. Using Monte Carlo Dropout~\cite{gal2016dropout}, we subject the decoder to dropout during $K$ stochastic forward passes on each sample, generating a series of coarse pseudo labels $\{\hat{Y}_{k}^{T}, k=1,...,K\}$. We compute the uncertainty map $m$ by taking the standard deviation $std(\hat{Y}_{1}^{T},...,\hat{Y}_{k}^{T})$ of the $K$ forward predictions. We then obtain a refined pseudo label $\acute{z}_{obj}^{T}$ by applying an uncertainty threshold $\eta$ to generate a stable prediction.

% However, the prototype of the target domain is unreliable due to false predictions of $G$. 
% Following~\cite{chen2021source}, we estimate the uncertainty to denoise the coarse pseudo labels. Based on Monte Carlo Dropout~\cite{gal2016dropout}, the decoder is in dropout condition and has $K$ times stochastic forward on one sample to obtain a series coarse pseudo labels $\{\hat{Y}_{k}^{T}, k=1,...,K\}$. 
% Then the uncertainty map $M$ is computed by the standard deviation $M = std(\hat{Y}_{1}^{T},...,\hat{Y}_{k}^{T})$ of the $K$ forward predictions. 
% Hence, the refined pseudo label $\acute{z}_{obj}^{T}$ can be obtained by configuring an uncertainty threshold $\eta$ to generate a stable prediction. 
	
% After refining the pseudo labels at the target domain, we further devise a prototype-guided consistency learning loss $\mathcal{L}_{pcl}$ to reduce the distance of the prototypes (i.e., $z_{obj}^{S}$ and $\acute{z}_{obj}^{T}$) from the source and target domains for domain adaptation.

After refining the pseudo labels in the target domain, we propose a prototype-guided consistency learning loss $\mathcal{L}_{pcl}$ to further narrow the gap between the source and target domains for domain adaptation. This loss aims to reduce the distance between the prototypes (i.e., $z_{obj}^{S}$ and $\acute{z}_{obj}^{T}$) from the source and target domains.
The definition of $\mathcal{L}_{pcl}$ is given by:
\begin{equation}\label{eq8}
		\mathcal{L}_{pcl} = || z_{obj}^{S}-\acute{z}_{obj}^{T}||,
	\end{equation}
By adding $\mathcal{L}_{pcl}$ with the supervised learning loss $\mathcal{L}_{seg}$ at the source domain, the total loss $\mathcal{L}_{M}$ of the stage 2 is computed by:
	\begin{equation}\label{eq9}
		\mathcal{L}_{M} = \mathcal{L}_{seg} + \lambda \mathcal{L}_{pcl}.
	\end{equation}
where the weight $\lambda$  is empirically set as  $\lambda=0.5$ in our experiments. 

\if 0
to courage the foregrounds of different domains to be closer and help the segmentor focus on the consistent content, the prototype-guided consistency learning loss $\mathcal{L}_{pcl}$ can be formulated as:
	\begin{equation}\label{eq8}
		\mathcal{L}_{pcl} = || z_{obj}^{S}-\acute{z}_{obj}^{T}||,
	\end{equation}
Combining $\mathcal{L}_{pcl}$ and the supervised learning loss $\mathcal{L}_{seg}$, the module loss $\mathcal{L}_{M}$ is computed by:
	\begin{equation}\label{eq9}
		\mathcal{L}_{M} = \mathcal{L}_{seg} + \lambda \mathcal{L}_{pcl}.
	\end{equation}
where $\lambda$  is a trade-off parameter, and its value is empirically set as $\lambda=0.5$ in our experiment. 
\fi

\section{Experiments}
\subsubsection{Dataset and evaluation metrics.}
We adopt three publicly available fundus datasets for our experiments, and each one is regarded as an identifiable domain showing a domain gap. 
In detail, we use the training set of the REFUGE~\cite{orlando2020refuge} as the source domain and test on the target domain RIM-ONE-r3~\cite{fumero2011rim} and Drishti-GS~\cite{sivaswamy2015comprehensive}. 
The source domain includes 400 annotated training images.
Following the data usage manner in~\cite{wang2019boundary}, we split 99/60 and 50/51 for training/testing in RIMONE-r3 and Drishti-GS respectively. 
We crop a 512$\times$512 disc region of interest (RoI) as the network input. The source domain has data augmentation operations, including random rotation, random flipping, elastic transformation, contrast adjustment,
adding Gaussian noise and random erasing, while we apply no operation for the target domain for UDA consideration. 
We introduce the Dice coefficient to compare different segmentation methods quantitatively.
%The segmentation metrics are verified by the Dice coefficient for pixel-wise accuracy measure.
\subsubsection{Implementation details.}
For Stage 1 of our method, the totally sampling timestep $t$ is $t=1000$ with $40k$ iterations. The optimizer adopts Adam with a learning rate of $1e-3$. And we utilize the representations from the middle blocks of U-Net decoder $B=\{5,6, 7, 8, 12\}$. The reverse diffusion process steps $t$ are $t=\{50,150,250\}$. As for Stage 2 of our method, in order to make a fair comparison with existing methods, we adopt the decoder of deeplabv3+~\cite{chen2018encoder} for the segmentor G. The dropout rate is $0.5$ for $M=8$ stochastic forward passes. The threshold $\gamma$ and $\eta$ are set as $0.75$ and $0.05$. The batch size of both is $8$, where $4/4$ is between source and target domains. 
We implement our network using Pytorch framework with $2$ RTX Titan Xp GPUs.

\subsection{Comparison with state-of-the-arts}

	\begin{table*}[!t]
	\centering
	
	\caption{Comparison of different approaches on the target domain datasets
	}
	\begin{center}
     \vspace{-3mm}
		\setlength\tabcolsep{0.5mm}
		\resizebox{0.95\textwidth}{!}{%
			\begin{tabular}{>{\centering}p{3.2cm}|>{\centering}p{2cm}|>{\centering}p{2cm}|>{\centering}p{2cm}|>{\centering\arraybackslash}p{2cm}}
				\Xhline{1.5pt}
				\multirow{2}{*}{Methods} &\multicolumn{2}{c|}{\textbf{RIM-ONE-r3}\cite{fumero2011rim}}&\multicolumn{2}{c}{\textbf{Drishti-GS}\cite{sivaswamy2015comprehensive}}\\
				\cline{2-5}
				
				{} &\textit{Dice disc} &\textit{Dice cup}  &\textit{Dice disc} &\textit{Dice cup}  \\
				
				\cline{1-5}
				Baseline(w/o DA) &\underline{0.946}&\underline{0.879}&\underline{0.974}&\underline{0.912}\\
				\cline{1-5}
				
				TD-GAN \cite{zhang2018task} &0.853&0.728&0.924&0.747\\
				\cline{1-5}
				Hoffman et al. \cite{hoffman2016fcns} &0.852&0.755&0.959&0.851\\
				\cline{1-5}
				Javanmardi et al. \cite{javanmardi2018domain} &0.853&0.779&0.961&0.849\\
				\cline{1-5}
				OSAL-pixel \cite{wang2019patch}&0.854&0.778&0.962&0.851\\
				\cline{1-5}
				pOSAL \cite{wang2019patch}&0.865&0.787&0.965&0.858\\
				\cline{1-5}
				BEAL \cite{wang2019boundary}&0.898&0.810&0.961&0.862\\
				\cline{1-5}
				%AdvEnt \cite{vu2019advent} &0.827&0.962&0.780&0.897\\
				\cline{1-5}
				CLR \cite{feng2022unsupervised}&0.905&0.841& \textbf{0.966}&\textbf{0.892}\\
				\cline{1-5}
				Ours &\textbf{0.913}&\textbf{0.852}&\textbf{0.966}&0.884\\
				\Xhline{1.5pt}

		\end{tabular}}
	\end{center}
	%\vspace{-10mm}
	\label{tab:1}
\end{table*}

\noindent
\textbf{Quantitative Comparison.} \ We compare our proposed method against seven state-of-the-art UDA algorithms and report their Dice scores for optic disc  and optic cup segmentation for Drishti-GS and RIM-ONE-r3 datasets in Table \ref{tab:1}.
Here, we first report a supervised baseline result (w/o DA) which takes vanilla DPM to extract features. Apparently, this result is the upper bounder of all UDA methods.
For the RIM-ONE-r3 dataset, CLR~\cite{feng2022unsupervised} has the best Dice score for optic disc and optic cup segmentation among all seven compared methods, and they are 0.905 and 0.841.
Contrarily, our method further outperforms CLR in terms of the optic disc and optic cup segmentation for the RIM-ONE-r3 dataset. It improves the Dice score from 0.905 to 0.913 for the optic disc segmentation, and the Dice score from 0.841 to 0.852 for the optic cup segmentation.
Regarding the Drishti-GS dataset, both our method and CLR~\cite{feng2022unsupervised} have the Dice score of 0.966, which is the largest one for the optic disc segmentation.
Although the Dice score (0.884) of our method for the optic cup segmentation takes the 2nd rank, it is only slightly smaller than the best one (0.892). 
We argue that such inferior performance (from 0.892 to 0.884) is caused by that BEAL~\cite{wang2019boundary} and CLR~\cite{feng2022unsupervised} adopt a pre-trained backbone from ImageNet, while our diffusion configuration only uses the fundus image data, which is pretty smaller than ImageNet.   

\if 0
%%
%Except for the dice coefficients (DI) of the optic cup on Drishti-GS, we have outperformed or equal to the state-of-the-art methods for the rest metrics. Compared to the best CLR~\cite{feng2022unsupervised}, we have equal DI of optic disc segmentation on Drishti-GS dataset while achieving $0.9\%$ and $0.8\%$ improvement for DI of the optic cup and optic disc on RIM-ONE-r3 dataset respectively. Noting that the deeplabv3+ based methods like BEAL~\cite{wang2019boundary} and CLR, adopt pre-trained backbone from ImageNet, while our diffusion configuration only uses the fundus image data. With less data access, our framework still performs powerfully suggesting that the DPM possesses potential in representation extraction. 
\fi

\begin{figure*}[!t]
		\centering
		\includegraphics[width=0.95\textwidth,]{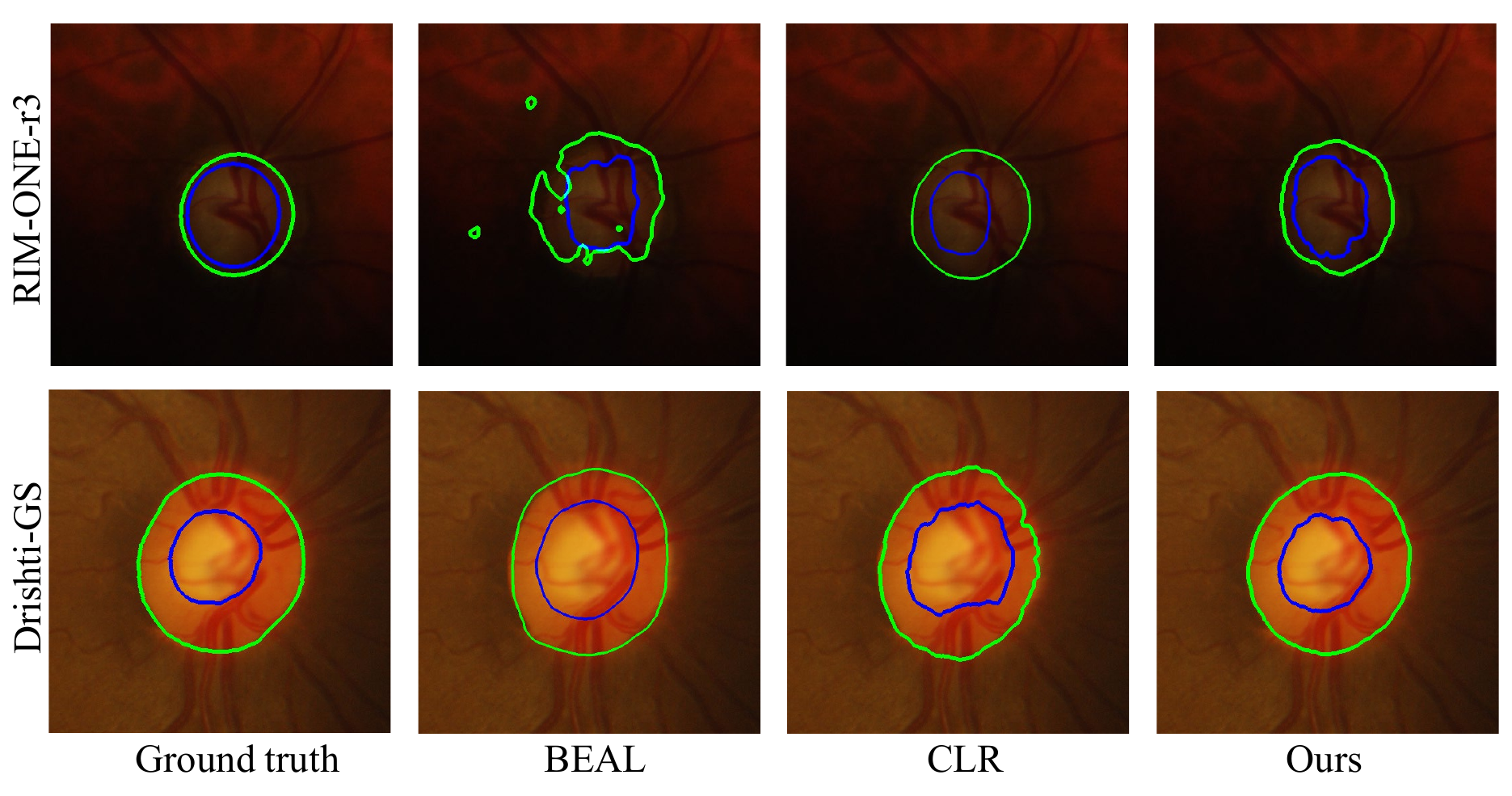}
            \vspace{-4mm}
		\caption{Visually comparing the optic disc and cup segmentation results produced by our network and state-of-the-art UDA methods. Apparently, our method can more accurately segment the optic disc and optic cup regions than compared methods.}
		\label{compare}
  \vspace{-6mm}
\end{figure*}

%\vspace{2mm}
\noindent
\textbf{Visual Comparison.} \ 
Fig. \ref{compare} visually compares the segmentation results produced by our network and state-of-the-art methods for the input images from Drishti-GS and RIM-ONE-r3. Apparently, our method can more accurately segment the optic disc (green) and optic cup (blue) regions, and our segmentation results are most consistent with the ground truth; see the first column of Fig. \ref{compare}. 
%We provide qualitative comparisons displayed in Fig. \ref{compare} as well. One can observe that our segmentation results are distributed more accurately, especially at the edges near the optic disc and optic cup.

\subsection{Ablation study}

\begin{table*}[t!]
		\centering
		\caption{Quantitative results of the ablation study experiment in our method. DR denotes the domain discriminator $\mathcal{L}_{D}$ of Eq.~\ref{eq6}.}
            \vspace{-3mm}
		\begin{tabular}{cccccc|ccc}
			\Xhline{1.5pt}
			\multicolumn{5}{l|}{\multirow{2}{*}{\quad  \quad \quad \quad \quad Configuration}}                                                               & \multicolumn{4}{c}{\quad Dataset}                                                                         \\ \cline{6-9} 
			
			\multicolumn{5}{l|}{}                                                                                      & \multicolumn{2}{c|}{\textbf{RIM-ONE-r3}\cite{fumero2011rim}}                                & \multicolumn{2}{c}{\textbf{Drishti-GS}\cite{sivaswamy2015comprehensive}}           \\ \hline
			\multicolumn{1}{c|}{Name}&\multicolumn{1}{c|}{Diff} & \multicolumn{1}{c|}{DA} & \multicolumn{1}{c|}{+DPL} & \multicolumn{1}{c|}{+PCL} & \multicolumn{1}{c|}{\textit{Dice disc}} & \multicolumn{1}{c|}{\textit{Dice cup}} & \multicolumn{1}{c|}{\textit{Dice disc}} & \textit{Dice cup} \\ \hline
			
			\multicolumn{1}{c|}{M1 (basic)}&\multicolumn{1}{c|}{\Checkmark}        & \multicolumn{1}{c|}{}         & \multicolumn{1}{c|}{}         &     \multicolumn{1}{c|}{}        & \multicolumn{1}{c|}{0.846}     & \multicolumn{1}{c|}{0.793}    & \multicolumn{1}{c|}{0.934}     & 0.809   \\ \hline

			%	\multicolumn{1}{c|}{\Checkmark}        & \multicolumn{1}{c|}{}         & \multicolumn{1}{c|}{\Checkmark}        &            & \multicolumn{1}{c|}{0.898}     & \multicolumn{1}{c|}{0.837}    & \multicolumn{1}{c|}{0.963}     & 0.875    \\ \hline
			
			%	\multicolumn{1}{c|}{\Checkmark}        & \multicolumn{1}{c|}{}         & \multicolumn{1}{c|}{}         & \Checkmark          & \multicolumn{1}{c|}{0.901}     & \multicolumn{1}{c|}{0.833}    & \multicolumn{1}{c|}{0.961}     & 0.881    \\ \hline
			
			\multicolumn{1}{c|}{M2}&\multicolumn{1}{c|}{\Checkmark}        & \multicolumn{1}{c|}{\Checkmark}        & \multicolumn{1}{c|}{}        &  \multicolumn{1}{c|}{}           & \multicolumn{1}{c|}{0.884}     & \multicolumn{1}{c|}{0.821}    & \multicolumn{1}{c|}{0.947}     & 0.869    \\ \hline
			
			\multicolumn{1}{c|}{M3}&\multicolumn{1}{c|}{\Checkmark}        & \multicolumn{1}{c|}{\Checkmark}        & \multicolumn{1}{c|}{\Checkmark}        & \multicolumn{1}{c|}{}            & \multicolumn{1}{c|}{\textbf{0.916}}     & \multicolumn{1}{c|}{0.829}    & \multicolumn{1}{c|}{0.963}     &0.876     \\ \hline
			
		\multicolumn{1}{c|}{M4 (Ours)}&	\multicolumn{1}{c|}{\Checkmark}        & \multicolumn{1}{c|}{\Checkmark}        & \multicolumn{1}{c|}{}        &  \multicolumn{1}{c|}{\Checkmark}        & \multicolumn{1}{c|}{0.913}     & \multicolumn{1}{c|}{\textbf{0.852}}    & \multicolumn{1}{c|}{\textbf{0.966}}     & \textbf{0.884}    \\ \Xhline{1.5pt}
		\end{tabular}
		\label{tab:2}
  
\end{table*}

We conduct ablation study experiments to investigate major components of our framework.
We first construct a baseline (denoted as ``M1'') by only extracting diffusion features to predict the segmentation results. It is equal to remove the domain discriminator $\mathcal{L}_{D}$ at Eq.~\ref{eq6} and the prototype-guided consistency loss $\mathcal{L}_{pcl}$ at Eq.~\ref{eq8} of the PCL module.
Then, we add the domain discriminator $\mathcal{L}_{D}$ into ``M1'' to construct another baseline network (``M2'').
After that, we add  a DPL~\cite{chen2021source} module into ``M2'' to build a baseline network (``M3''), and the prototype-guided consistency loss $\mathcal{L}_{pcl}$ into ``M2'' to build ``M4''. Apparently, ``M4'' is equal to the full setting of our network. 

Table \ref{tab:2} reports the Dice score of the optic disc and cup segmentation of our method and baseline networks in terms of Drishti-GS and RIM-ONE-r3 datasets.
Apparently, ``M2'' has larger Dice scores than ``M1'' on the optic disc and cup segmentation for both two datasets. It indicates that the domain discriminator $\mathcal{L}_{D}$ of Eq.~\ref{eq6} on diffusion features can reduce the domain gap of the source and target domains, thereby improving the UDA performance.
Then,  ``M3'' and our method (i.e., ``M4'') outperform ``M2'' for the optic disc and cup segmentation on the two datasets.
It demonstrates that exploring the prototype information to compute consistency loss can further enhance the domain adaptation performance.
More importantly, our method has a superior Dice score over ``M3''. It indicates that our PCL can better reduce the domain gap than DPL~\cite{chen2021source}.
Specifically, compared to ``M3'', our method improves the Dice score of the optic disc segmentation from 0.963 to 0.966, and the Dice score of the optic cup segmentation from 0.876 to 0.844 on the Drishti-GS dataset.
For the RIM-ONE-r3 dataset, our method improves the Dice score of the optic cup segmentation from 0.829 to 0.852. And the Dice scores of the optic disc segmentation for our method and ``M3'' are very close, and they are 0.916 and 0.913, respectively.

\if 0
In Table \ref{tab:2}, the Diff (Baseline) and DADiff indicate the vanilla diffusion and our proposed one. With a generalized feature provided by DADiff, there is a noticeable improvement compared to the one without equipped. We also conduct pseudo-labeling method DPL~\cite{chen2021source} in our experiments since it also utilizes prototype but in an offline way to denoise the pseudo label, while we alleviate the prototype discrepancy via PCL online. It can be seen that both DPL and PCL help to boost performance, while the latter prevails. This is because the prototypes from the target domain are dynamically adjusted and aggregated as the training progresses, reducing the distance between the target and the source domain such that the segmentor enables to learn the consistent objectives.
\fi

\section{Conclusion}
% In this paper, we develop a Diffusion-based and Prototype-guided network (DP-Net) for Unsupervised Domain Adaptive Segmentation. On the one hand, we explore the potential of the diffusion model in domain adaptation. The proposed Distribution Aligned Diffusion (DADiff) is used to extract the generalized features. On the other, we design Prototype-guided Consistency Learning (PCL) module using the foreground prototype to instruct the segmentor to learn consistent objectives. Experimental results on benchmark datasets show that our network outperforms state-of-the-art methods, suggesting it can be effective for other unsupervised domain adaptation tasks.
In this paper, we propose the Diffusion-based and Prototype-guided network (DP-Net) for Unsupervised Domain Adaptive Segmentation. Specifically, we investigate the effectiveness of the diffusion model for domain adaptation, and propose the Distribution Aligned Diffusion (DADiff) approach to extract generalized features. Additionally, we develop the Prototype-guided Consistency Learning (PCL) module that utilizes the foreground prototype to guide the segmentor in learning consistent objectives. Our experimental results on benchmark datasets demonstrate the superior performance of DP-Net over state-of-the-art methods, indicating its potential for other unsupervised domain adaptation tasks.
	
\bibliographystyle{splncs04.bst}
	
\bibliography{paper1883}

\begin{thebibliography}{10}
\providecommand{\url}[1]{\texttt{#1}}
\providecommand{\urlprefix}{URL }
\providecommand{\doi}[1]{https://doi.org/#1}

\bibitem{amit2021segdiff}
Amit, T., Nachmani, E., Shaharbany, T., Wolf, L.: Segdiff: Image segmentation
  with diffusion probabilistic models. arXiv preprint arXiv:2112.00390  (2021)

\bibitem{baranchuk2021label}
Baranchuk, D., Rubachev, I., Voynov, A., Khrulkov, V., Babenko, A.:
  Label-efficient semantic segmentation with diffusion models. arXiv preprint
  arXiv:2112.03126  (2021)

\bibitem{bateson2020source}
Bateson, M., Kervadec, H., Dolz, J., Lombaert, H., Ben~Ayed, I.: Source-relaxed
  domain adaptation for image segmentation. In: Medical Image Computing and
  Computer Assisted Intervention--MICCAI 2020: 23rd International Conference,
  Lima, Peru, October 4--8, 2020, Proceedings, Part I 23. pp. 490--499.
  Springer (2020)

\bibitem{chen2021source}
Chen, C., Liu, Q., Jin, Y., Dou, Q., Heng, P.A.: Source-free domain adaptive
  fundus image segmentation with denoised pseudo-labeling. In: Medical Image
  Computing and Computer Assisted Intervention--MICCAI 2021: 24th International
  Conference, Strasbourg, France, September 27--October 1, 2021, Proceedings,
  Part V 24. pp. 225--235. Springer (2021)

\bibitem{chen2018encoder}
Chen, L.C., Zhu, Y., Papandreou, G., Schroff, F., Adam, H.: Encoder-decoder
  with atrous separable convolution for semantic image segmentation. In:
  Proceedings of the European conference on computer vision (ECCV). pp.
  801--818 (2018)

\bibitem{chen2022diffusiondet}
Chen, S., Sun, P., Song, Y., Luo, P.: Diffusiondet: Diffusion model for object
  detection. arXiv preprint arXiv:2211.09788  (2022)

\bibitem{dhariwal2021diffusion}
Dhariwal, P., Nichol, A.: Diffusion models beat gans on image synthesis.
  Advances in Neural Information Processing Systems  \textbf{34},  8780--8794
  (2021)

\bibitem{feng2022unsupervised}
Feng, W., Wang, L., Ju, L., Zhao, X., Wang, X., Shi, X., Ge, Z.: Unsupervised
  domain adaptive fundus image segmentation with category-level regularization.
  In: Medical Image Computing and Computer Assisted Intervention--MICCAI 2022:
  25th International Conference, Singapore, September 18--22, 2022,
  Proceedings, Part II. pp. 497--506. Springer (2022)

\bibitem{fumero2011rim}
Fumero, F., Alay{\'o}n, S., Sanchez, J.L., Sigut, J., Gonzalez-Hernandez, M.:
  Rim-one: An open retinal image database for optic nerve evaluation. In:
  international symposium on computer-based medical systems. pp.~1--6. IEEE
  (2011)

\bibitem{gal2016dropout}
Gal, Y., Ghahramani, Z.: Dropout as a bayesian approximation: Representing
  model uncertainty in deep learning. In: international conference on machine
  learning. pp. 1050--1059. PMLR (2016)

\bibitem{ganin2015unsupervised}
Ganin, Y., Lempitsky, V.: Unsupervised domain adaptation by backpropagation.
  In: International conference on machine learning. pp. 1180--1189. PMLR (2015)

\bibitem{ho2020denoising}
Ho, J., Jain, A., Abbeel, P.: Denoising diffusion probabilistic models.
  Advances in Neural Information Processing Systems  \textbf{33},  6840--6851
  (2020)

\bibitem{hoffman2016fcns}
Hoffman, J., Wang, D., Yu, F., Darrell, T.: Fcns in the wild: Pixel-level
  adversarial and constraint-based adaptation. arXiv preprint arXiv:1612.02649
  (2016)

\bibitem{hsu2020every}
Hsu, C.C., Tsai, Y.H., Lin, Y.Y., Yang, M.H.: Every pixel matters: Center-aware
  feature alignment for domain adaptive object detector. In: Computer
  Vision--ECCV 2020: 16th European Conference, Glasgow, UK, August 23--28,
  2020, Proceedings, Part IX 16. pp. 733--748. Springer (2020)

\bibitem{javanmardi2018domain}
Javanmardi, M., Tasdizen, T.: Domain adaptation for biomedical image
  segmentation using adversarial training. In: 2018 IEEE 15th International
  Symposium on Biomedical Imaging (ISBI 2018). pp. 554--558. IEEE (2018)

\bibitem{moreno2012unifying}
Moreno-Torres, J.G., Raeder, T., Alaiz-Rodr{\'\i}guez, R., Chawla, N.V.,
  Herrera, F.: A unifying view on dataset shift in classification. Pattern
  recognition  \textbf{45}(1),  521--530 (2012)

\bibitem{orlando2020refuge}
Orlando, J.I., Fu, H., Breda, J.B., van Keer, K., Bathula, D.R., et~al.: Refuge
  challenge: A unified framework for evaluating automated methods for glaucoma
  assessment from fundus photographs. Medical image analysis  \textbf{59},
  101570 (2020)

\bibitem{shaban2019staingan}
Shaban, M.T., Baur, C., Navab, N., Albarqouni, S.: Staingan: Stain style
  transfer for digital histological images. In: 2019 Ieee 16th international
  symposium on biomedical imaging (Isbi 2019). pp. 953--956. IEEE (2019)

\bibitem{sivaswamy2015comprehensive}
Sivaswamy, J., Krishnadas, S., Chakravarty, A., Joshi, G., Tabish, A.S.,
  et~al.: A comprehensive retinal image dataset for the assessment of glaucoma
  from the optic nerve head analysis. JSM Biomedical Imaging Data Papers
  \textbf{2}(1), ~1004 (2015)

\bibitem{sun2022safe}
Sun, T., Lu, C., Zhang, T., Ling, H.: Safe self-refinement for
  transformer-based domain adaptation. In: Proceedings of the IEEE/CVF
  Conference on Computer Vision and Pattern Recognition. pp. 7191--7200 (2022)

\bibitem{tritrong2021repurposing}
Tritrong, N., Rewatbowornwong, P., Suwajanakorn, S.: Repurposing gans for
  one-shot semantic part segmentation. In: Proceedings of the IEEE/CVF
  conference on computer vision and pattern recognition. pp. 4475--4485 (2021)

\bibitem{wang2019boundary}
Wang, S., Yu, L., Li, K., Yang, X., Fu, C.W., Heng, P.A.: Boundary and
  entropy-driven adversarial learning for fundus image segmentation. In:
  Medical Image Computing and Computer Assisted Intervention--MICCAI 2019: 22nd
  International Conference, Shenzhen, China, October 13--17, 2019, Proceedings,
  Part I 22. pp. 102--110. Springer (2019)

\bibitem{wang2019patch}
Wang, S., Yu, L., Yang, X., Fu, C.W., Heng, P.A.: Patch-based output space
  adversarial learning for joint optic disc and cup segmentation. IEEE
  transactions on medical imaging  \textbf{38}(11),  2485--2495 (2019)

\bibitem{xu2021generative}
Xu, Y., Shen, Y., Zhu, J., Yang, C., Zhou, B.: Generative hierarchical features
  from synthesizing images. In: Proceedings of the IEEE/CVF Conference on
  Computer Vision and Pattern Recognition. pp. 4432--4442 (2021)

\bibitem{yan2022unsupervised}
Yan, P., Wu, Z., Liu, M., Zeng, K., Lin, L., Li, G.: Unsupervised domain
  adaptive salient object detection through uncertainty-aware pseudo-label
  learning. In: Proceedings of the AAAI Conference on Artificial Intelligence.
  vol.~36, pp. 3000--3008 (2022)

\bibitem{yang2023tvt}
Yang, J., Liu, J., Xu, N., Huang, J.: Tvt: Transferable vision transformer for
  unsupervised domain adaptation. In: Proceedings of the IEEE/CVF Winter
  Conference on Applications of Computer Vision. pp. 520--530 (2023)

\bibitem{zhang2021prototypical}
Zhang, P., Zhang, B., Zhang, T., Chen, D., Wang, Y., Wen, F.: Prototypical
  pseudo label denoising and target structure learning for domain adaptive
  semantic segmentation. In: Proceedings of the IEEE/CVF conference on computer
  vision and pattern recognition. pp. 12414--12424 (2021)

\bibitem{zhang2018task}
Zhang, Y., Miao, S., Mansi, T., Liao, R.: Task driven generative modeling for
  unsupervised domain adaptation: Application to x-ray image segmentation. In:
  Medical Image Computing and Computer Assisted Intervention--MICCAI 2018: 21st
  International Conference, Granada, Spain, September 16-20, 2018, Proceedings,
  Part II. pp. 599--607. Springer (2018)

\bibitem{zheng2021rectifying}
Zheng, Z., Yang, Y.: Rectifying pseudo label learning via uncertainty
  estimation for domain adaptive semantic segmentation. International Journal
  of Computer Vision  \textbf{129}(4),  1106--1120 (2021)

\end{thebibliography}
\end{document}